\documentclass{bmvc2k}

\usepackage{amsmath}
\usepackage{multirow}
\usepackage{adjustbox}
\usepackage{graphicx}
\usepackage{amsfonts}
\usepackage{wrapfig,lipsum,booktabs}
\usepackage[svgnames]{xcolor}
\usepackage{threeparttable}
\usepackage{graphicx}
\title{Group Based Deep Shared Feature Learning for Fine-grained Image Classification}

\addauthor{Xuelu Li}{xul76@psu.edu}{1}
\addauthor{Vishal Monga}{vmonga@engr.psu.edu }{1}

\addinstitution{
 Department of Electrical Engineering\\
 Pennsylvania State University\\
 State College, U.S.
}

\runninghead{Li \protect\etal}{Group Based Deep Shared Feature Learning}

\def\etal{\emph{et al}\bmvaOneDot}

\newcommand{\bX}{\mathbf{X}}

\newcommand{\by}{\mathbf{y}}

\newcommand{\bI}{\mathbf{I}}

\newcommand{\bl}{\mathbf{l}}

\newcommand{\bm}{\mathbf{m}}
\newcommand{\bs}{\mathbf{s}}
\newcommand\numberthis{\addtocounter{equation}{1}\tag{\theequation}}
\newcommand{\norm}[1]{||#1||}

\begin{document}
\setlength{\abovedisplayskip}{5pt}
\setlength{\belowdisplayskip}{5pt}
\setlength{\textfloatsep}{0.1cm}
\maketitle

\begin{abstract}	
Fine-grained image classification has emerged as a significant challenge because objects in such images have small inter-class visual differences but with large variations in pose, lighting, and viewpoints, etc. Most existing work focuses on highly customized feature extraction via deep network architectures which have been shown to deliver state of the art performance. Given that images from distinct classes in fine-grained classification share significant features of interest, we present a new deep network architecture that explicitly models shared features and removes their effect to achieve enhanced classification results. Our modeling of shared features is based on a new group based learning wherein existing classes are divided into groups and multiple shared feature patterns are discovered (learned). We call this framework Group based deep Shared Feature Learning (GSFL) and the resulting learned network as GSFL-Net. Specifically, the proposed GSFL-Net develops a specially designed autoencoder which is constrained by a newly proposed Feature Expression Loss to decompose a set of features into their constituent shared and discriminative components. During inference, only the discriminative feature component is used to accomplish the classification task. A key benefit of our specialized autoencoder is that it is versatile and can be combined with state-of-the-art fine-grained feature extraction models and trained together with them to improve their performance directly. Experiments on benchmark datasets show that GSFL-Net can enhance classification accuracy over the state of the art with a more interpretable architecture.
\end{abstract}

%-------------------------------------------------------------------------
\section{Introduction}
\label{sec:intro}
%\vspace{-5pt}
Fine-grained recognition involves classification of instances within a subordinate category, such as species of birds, flowers, and animals, models of cars and makes of airplanes, etc \cite{wah2011caltech,maji2013fine,krause20133d}. Different from conventional image classification, the tasks of fine-grained recognition are required to discriminate highly localized attributes of objects while being invariant to their poses and locations in the images since objects in such images always contain higher visual similarity and are surrounded by various complex environments. Furthermore, due to the difficulty of obtaining examples for different classes, training imagery is often limited in practice.

Considering that discriminative attributes are typically located in parts of the objects, most prior work for fine-grained image classification focuses on handling the variations in lighting, viewpoints, and poses using localization techniques \cite{lin2015bilinear,jaderberg2015spatial,zhang2016weakly,krause2015fine,zhang2015fine}. There is also some work that tries to train datasets with additional data from the web \cite{krause2016unreasonable,cui2016fine} to enhance performance in the face of limited training. 

Although much effort has focused on extracting discriminative attributes of objects, there has also been work that addresses the issue of inter-class visual similarity. We note that this issue brings both challenges and benefits. The attention to local attributes of objects for fine-grained classification is in fact inspired from the same observation but this technique often requires excessive customization for datasets at hand and limits the overall applicability of the techniques. On the other hand, a benefit is the understanding that if similar or shared features across classes are accounted for in the classification process, the difficulty of fine-grained classification can be dramatically reduced. In Sparse Representation based Classification (SRC), a similar idea has been explored. For example, in \cite{zhang2017fine}, the authors try to separate fine-grained images by jointly learning the encoding parameters and codebooks through low-rank sparse coding with general and class specific codebook generation. However, restricted to the nature of SRC based methods, the performance of the method is limited when applied to datasets with a large number of classes. Besides, in \cite{zhang2017fine}, only one general code book is learned based on the assumption that all classes share common features. However, when dealing with datasets with a large number of classes, it is more reasonable that some clusters of classes be organized as groups. 

Deep learning has recently emerged to supplant the state of the art in image classification \cite{lecun2015deep,reed2016learning,xie2015hyper,zhou2016learning,chen2015deep,akata2015evaluation,liu2015deep,donahue2014decaf,zhang2016embedding,oh2016deep}. A more detailed discussion of deep architectures for fine grained image classification is provided in Section \ref{sec:related}.
Inspired by the aforementioned discussion, we propose a Group based Deep Shared Feature Learning Network (GSFL-Net) that can extract shared as well as discriminative features for fine-grained classification. Our modeling of shared features is based on a new group based learning wherein existing classes are divided into groups and multiple shared feature patterns are discovered (learned). 
\begin{wrapfigure}{r}{5.5cm}
%	\vspace{-10pt}
	\includegraphics[width=0.44\textwidth]{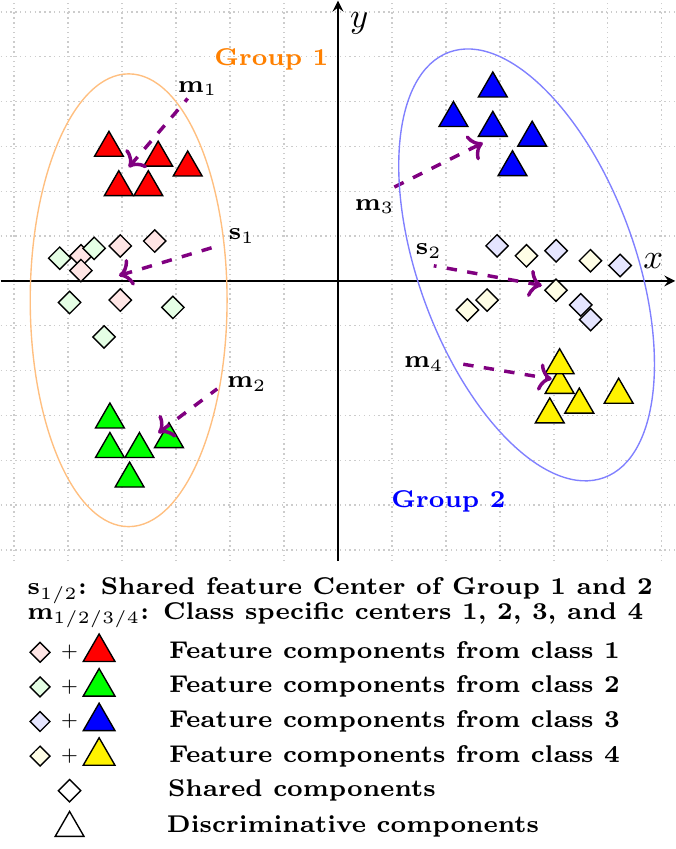}
	\caption{Motivation for GSFL-Net.}
	\label{fig:motivation}
%	\vspace{-10pt}
\end{wrapfigure}
The {\bf motivation} behind GSFL-Net is shown in Fig. \ref{fig:motivation}: The classes are divided into groups
according to the distances between each class specific feature center ($\bm_i$) in the feature space. In particular, Fig. \ref{fig:motivation} illustrates 4 classes and 2 groups. The main idea is that by removing the effects brought by the shared feature patterns within each group, the rest of the discriminative components can be quite effective for fine-grained classification. Note that Fig.\ \ref{fig:motivation} additionally shows two different kinds of feature centers: labeled $\bm_1$ through $\bm_4$ and $\bs_1$ through $\bs_2$. 
$\bm_i$ represents the class-specific feature center, which is computed from the discriminative feature components from the $i$th class. $\bs_j$ is the shared feature center for the $j$th group.  The proposed GSFL-Net is then constructed based on the following key observation:  The principal difficulty in classification of image datasets with a large number of classes is because a subset of those classes that contain highly similar images are particularly hard to discriminate between. This challenge can be addressed by designing encoders (see Fig. \ref{fig:NetworkArchitecture}) that decompose a set of features into shared and discriminative components in a way that grouping of classes is permitted and exploited. The feature centers shown in Fig.\ \ref{fig:motivation} are explicitly used in building regularization terms which guide the training process (see Section \ref{subsec:lossfunction}) and help learn powerful shared and discriminative feature components. In a departure from past efforts in modeling shared features \cite{zhang2017fine}, in our group based framework, there is no need to learn the shared feature center among features from classes which are already sufficiently different.

The proposed GSFL-Net is compared against state of the art in fine-grained classification on three benchmark datasets. Performance gains are shown for all three datasets along with versatility benefits in that GSFL-Net architecture can incorporate and be trained together with best-known feature extractors \cite{simonyan2014very,lin2018bilinear}.

%While many existing complex feature extraction models can only be applied to a specific image dataset, an obvious benefit of GSFL-Net is that it can be designed with a variety of powerful feature extractors by adding a simple autoencoder structure to the output of the feature extraction model (see Fig. ...) and be trained together with the feature extraction model in an end-to-end manner. This makes our network have the ability to improve the performance of the state-of-the-art feature extraction models that have been well applied to fine-grained image classification directly. Our {\bf contributions} in this paper mainly include: {\bf 1.} Propose the novel idea of optimizing feature learning for large class datasets by removing the shared feature patterns within each class group. {\bf 2.} Design a novel network architecture and its loss function. {\bf 3.} Broad experimental validation and insights.
%\vspace{-15pt}
\section{Related Work}
\label{sec:related}
%\vspace{-8pt}
Fine-grained image classification has been researched actively by the design of deep networks for highly localized and discriminative feature extraction. An approach that localizes the discriminative attributes was proposed in \cite{zheng2017learning}, where the authors develop a  part learning approach via a multi-attention convolutional neural network (CNN).  In \cite{yao2018autobd}, the authors propose a robust and discriminative visual descriptor by combining two complementary part-level and object-level visual descriptions. Another approach along the same lines is \cite{peng2018object}, where object-level and part-level `attentions' are jointly employed towards learning of multi-view and multi-scale features.

There is also research with the flavor of extracting more representative features by combing multiple network structures. For example, the authors in \cite{ge2015subset} propose a deep network system to partition images into $K$ subsets of similar images and learns an expert DCNN for each subset, then the output from each of the $K$ DCNNs is combined to form a single classification decision. In \cite{lin2018bilinear}, Lin, T. et al. present a simple network which can represent an image as a pooled outer product of features derived from two CNNs and capture localized feature interactions in a translation invariant manner. Shu, K., et at \cite{kong2017low} propose to represent the covariance features as a matrix and apply a low-rank bilinear classifier to address the computational demands of high feature dimensionality. Du, A., et al. \cite{dubey2018pairwise} propose a novel optimization procedure for end-to-end neural network training to reduce over-fitting. Gao proposes two compact bilinear representations in \cite{gao2016compact} with the same discriminative power as the full bilinear representation but with the benefits of significantly reduced feature dimension. Cui \cite{cui2017kernel} propose a general pooling framework that captures higher order feature interactions with the help of kernels. Similar work can also be found in \cite{yu2018hierarchical,wei2018grassmann}. 
%\vspace{-14pt}
\section{Group Based Deep Shared Feature Learning Network}
%------------------------------------------------------------------------- 
%\vspace{-9pt}
\subsection{Overview of the central idea}
\label{subsec:overview}

The classification task is essentially a process of forming the boundaries among different classes in the feature space. The wrongly classified feature samples are those that reside at the wrong sides of the boundaries. Such feature samples invariably have some shared components with those from the class which they are wrongly classified to. If we decompose the extracted features into shared and discriminative (class-specific) components, and subsequently base the classification decision only on the discriminative components -- then the accuracy of fine-grained classification can be significantly improved. 

For simplicity, imagine a two-class scenario. It is a reasonable assumption that the shared components should be concentrated somewhere near the middle of the "line" that connects class specific feature centers. The discriminative components, on the other hand, are concentrated near the (corresponding) class specific feature center. This is visually illustrated in Fig.\  \ref{fig:motivation}. The same observation also extends when there are multiple classes except that some classes may have stronger similarity than others and hence in identifying shared features, a grouping of such classes is additionally desired.

Based on the aforementioned observations, the goal of the proposed GSFL-Net is to decompose features into shared and discriminative components. This discovery of shared and discriminative components is achieved by designing a custom design autoencoder that employs two encoders (resp. for shared and discriminative components) and one decoder. Further in the learning, the classification loss is regularized by a novel feature expression loss term that enables a group based extraction of shared feature components by forcing closeness to a shared feature center for that group. Similarly, discriminative components are encouraged via a term within the feature expression loss that forces closeness to a class-specific feature center. The shared and class-specific feature centers are {\em not fixed} but also updated as a part of the learning procedure.

\begin{figure}[!tbp]
	\centering
	\centering
	\includegraphics[width=1\textwidth]{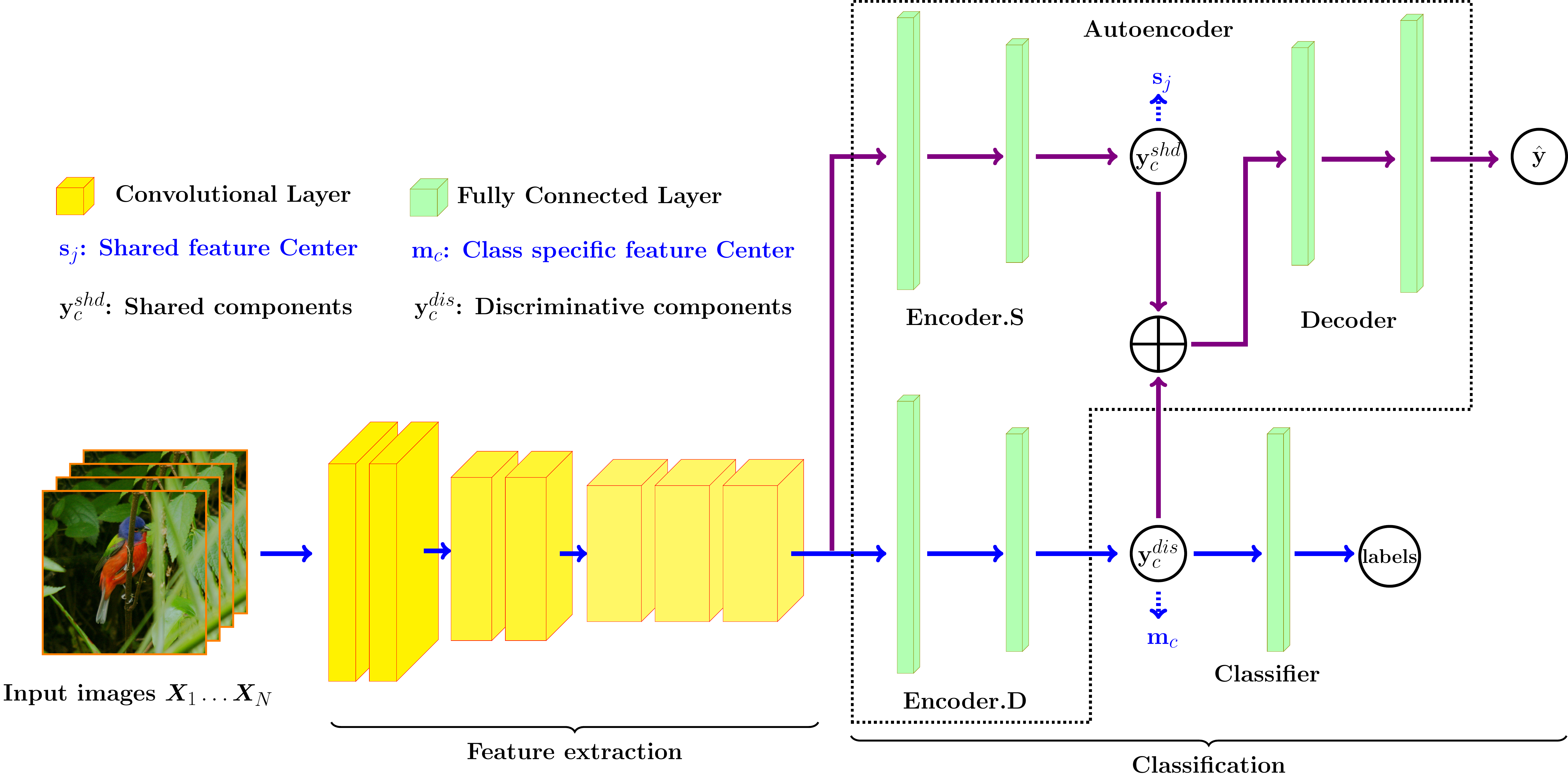}
	\caption{The network architecture of GSFL-Net: The network is composed of the Feature extraction part and the Classification part. The Feature Extraction part is used to extract the features from the input images and the Classification part is use to decompose the extracted features into shared features and discriminative features and finish the classification job. The whole network is trained from end to end. Note that we omit the activation layers between each layer in the figure. Details of the network architecture can be inferred from our implementation code}\protect\footnotemark[1].
	\label{fig:NetworkArchitecture}
%	\vspace{-5pt}
\end{figure} 
\footnotetext[1]{The code is available at \url{https://github.com/xueluli/GSFL-Net}}
%\vspace{-12pt}
\subsection{Network architecture}
%\vspace{-3pt}
The proposed network architecture is shown in Fig. \ref{fig:NetworkArchitecture} and consists of two different parts: the Feature extraction and the Classification part. The feature extraction part includes CNN layers which are used to extract representative features from the images. The feature extraction layers in Fig. \ref{fig:NetworkArchitecture} (in yellow) can in fact mimic state of the art known feature extraction methods  \cite{lin2018bilinear,gao2016compact,cui2017kernel}. Our contribution is then in feature decomposition and classification to realize the goals mentioned in Section \ref{subsec:overview}. This part consists of an autoencoder and an FC layer
%\begin{wraptable}{r}{4.5cm}
%	\vspace{-10pt}
%	\caption{The network structure of the classification part \big(Layer (input size, output size)\big)} % title of Table
%	\centering % used for centering table
%	\begin{adjustbox}{width=0.35\textwidth}
%		\scalebox{0.6}{
%			\begin{tabular}{c|c} % centered columns (4 columns)
%				\hline\hline %inserts double horizontal lines			
%				Encoder.S  & Encoder.D \tnote{a}\\
%				\hline
%				Dropout ($p=0.5$) \tnote{a}& Dropout ($p=0.5$)\\
%				FC \big(dim($\by$), 4096\big) & FC \big(dim($\by$\big), 4096) \\
%				Dropout ($p=0.5$) &  Dropout ($p=0.5$)\\
%				FC (4096, 4096)&FC(4096, 4096) \\
%				Relu &  Relu \\
%				\hline
%				Decoder& Classifier\\
%				\hline
%				FC (4096, 4096)& FC (4096, C)\\
%				Relu & Softmax\\
%				FC\big(4096, dim($\by$)\big)&\\
%				Sigmoid & \\
%				\hline %inserts single line
%			\end{tabular}}
%		\end{adjustbox}
%		\begin{tablenotes}
%			\item[a] \small $p$ is the dropping out rate, dim($\by$) is dimension of $\by$.
%		\end{tablenotes}
%		\label{table:AutoEncoder} % is used to refer this table in the text
%		\vspace{-10pt}
%	\end{wraptable}
	as the classifier. Different from other typical autoencoders, the proposed autoencoder includes two encoders, denoted as Encoder.S and Encoder.D, respectively, and one decoder, denoted as Decoder. The two encoders are used to decompose the feature $\by$ extracted from the feature extraction part into its  shared $\by_c^{shd}$ (output of Encoder.S) and the discriminative components $\by_c^{dis}$ (output of Encoder.D) -- respectively. The decoder is utilized to ensure that the sum of $\by_c^{shd}$ and $\by_c^{dis}$ can be reconstructed as $\by$ after sent through the Decoder (namely, the output of Decoder $\hat{\by}$ approximately equals to $\by$). This avoids the loss of essential information during the decomposition. Note that only $\by_c^{dis}$ will be used as the input of the following FC layer. Combined with the following Softmax layer, the FC layer accomplishes the classification task. It is worth noting that, after the training process, only the structures connected by the blue arrows as shown in Fig. \ref{fig:NetworkArchitecture} are saved for testing new images, which is memory efficient.
	%------------------------------------------------------------------------- 
%	\vspace{-10pt}
	\subsection{Regularized Loss function}\label{subsec:lossfunction}
%	\vspace{-3pt}
	To ensure that every part of the network architecture shown in Fig.\ref{fig:NetworkArchitecture} is able to learn outputs as expected, we also propose a novel regularized loss function to help restrict the behavior of the network. The proposed loss function mainly consists of Classification Loss and a newly proposed Feature Expression Loss (the weight decay loss is not detailed here since it is commonly used in deep learning based methods.) as below:
	
	1) {\bf Classification Loss}: The ultimate goal of our proposed network is classification. To this end, the standard cross entropy loss that is commonly used in the deep learning based classification frameworks \cite{lecun2015deep} is applied; we call it Classification Loss, and denote it as $L_{CF}$:
%	\vspace{-10pt}
	\begin{equation}
	L_{CF}=-\frac{1}{N}\sum_{i=1}^{N}\langle \bl_n, \log G(\bX_n)\rangle
	\end{equation}
%	\vspace{-1pt}  
	where $\bX=\{\bX_1,\cdots,\bX_n,\cdots,\bX_N\}$ represents the batch of input images, and $\bX_n$ represents the $n^{th}$ input image. $\bl_n$ represents the true label of $\bX_n$, and $N$ represents the total number of input images. $G(\bX_n)$ represents the output of the Softmax layer at the end of the classifier with respect to $\bX_n$. $\log(\cdot)$ represents the natural logarithm and $\langle \cdot \rangle$ represents the inner product. 
	%The cost function mainly plays its role between feature extraction part, Encoder.D and the classifier to finish the classification work.
	
	2) {\bf Feature Expression Loss}:  The loss is denoted as $L_{FE}$ and expressed as below:
	\begin{equation}
	\label{eq:FE}
	L_{FE}
	=\alpha_1\underbrace{\norm{\by-\hat{\by}}^2_F}_{\text{reconstruction loss}}+\alpha_2\underbrace{\sum_{c=1}^C\norm{\by^{\text{dis}}_c-\bm_c}^2_F}_{\text{class specific feature loss}}+\alpha_3\underbrace{{\sum_{j=1}^{N_g}\sum_{c\in\bI_j}^C\norm{\by^{\text{shd}}_c-\bs_j}^2_F}}_{\text{shared feature loss}}
	\end{equation}  
	where $\alpha_1$, $\alpha_2$, and $\alpha_3$ are hyperparameters balancing the weights of different regularizations, $\bm_c$ is $c^{th}$ class specific feature center and $\bs_j$ is the shared feature center of group $I_j$, $C$ is the number of classes, and $N_g$ is the number of groups and chosen by cross-validation \cite{monga2017handbook} (eg: as shown in Fig. \ref{fig:motivation}, $N_g=2$, $C=4$). In practice, before learning of the network, an unsupervised clustering technique is used to assign classes to $N_g$ groups and hence the membership of a given class to a group is known -- see Section \ref{subsec:updataM} for more details.
	
	From $L_{FE}$, it may be observed that the Feature Expression Loss consists of three complementary terms: The first term is the reconstruction loss, which is used to make sure that the sum of the outputs from the Encoder.S and the Encoder.D, $\by_c^{shd}$ and $\by_c^{dis}$, can be approximately reconstructed as the feature $\by$ through Decoder, whose output is $\hat{\by}$. The second term is the class specific feature loss, which is used to help the learned discriminative components $\by_c^{dis}$ to be close to the class specific feature center $\bm_c$. This ensures that there will be less misclassified samples around the boundaries between different classes. The last term is the shared feature loss, which encourages the learned shared components $\by_c^{shd}$ to concentrate around the shared feature center $\bs_j$. Through the Feature Expression Loss, we can separate $\by_c^{shd}$ from $\by_c^{dis}$, and utilize $\by_c^{dis}$ to accomplish the classification task. 
	
	The overall loss function $L$ of GSFL-Net is hence given by $L=L_{CF}+L_{FE}$.
%	\vspace{-10pt}
	%------------------------------------------------------------------------
	\subsection{ Group formation and strategy for updating $\bm_c$ and $\bs_j$}
	\label{subsec:updataM}
%	\vspace{-6pt}
	A key aspect of our design is learning the values of the class specific feature center $\bm_c$ and the group based shared feature center $\bs_j$ in Eq.\eqref{eq:FE}. Ideally, they should be updated by taking the entire training set into account and averaging the feature values of every class in each iteration. However, this is inefficient and even impractical in deep learning. Inspired by \cite{wen2016discriminative}, we perform the updates of the feature centers based on mini-batch instead of updating the centers with respect to the entire training set. In each iteration, the class specific center $\bm_c$ is computed by averaging $\by_c^{shd}$s in that batch, namely, $\bm_c$ may not be updated if no $\by_c^{shd}$ is available. Then, to avoid the large perturbations caused by a few mislabeled samples, a scalar $\eta$ is used to control the learning rate of the centers. Thus, the values of $\bm_c$ can be updated as below:
	\begin{align*}
		\Delta \bm_c=\frac{\sum_{i=1}^{N}\delta(l_i=c)(\bm_c-\by^{\text{dis}}_i)}{1+\sum_{i=1}^{N}\delta(l_i=c)}\numberthis
	\end{align*}
	\begin{align*}
		\bm_c = \bm_c-\eta\Delta \bm_c\numberthis
	\end{align*}
	where $\delta(\cdot)=1$ if $l_i=c$, equals $0$ if not.
	
	As the group based shared feature center, the value of $\bs_j$ is restricted by features from each class in $j^{th}$ group, and change according to the variations of class specific feature centers in each iteration. As a result, we need to learn multiple different shared feature centers for all the shared groups. The exact procedure is as follows:
	{\bf 1.} Use a pre-trained network (VGG 16 in our example) to extract features from each image.  
	{\bf 2.} Cluster the features into several different groups (for a given value of $N_g$) using a competitive unsupervised algorithm -- we employed the popular and simple $k$-means clustering \cite{pollard1981strong}. We say each class belongs to the group where most of the feature samples in the class are clustered to.
	{\bf 3.} Update the shared feature center $\bs_j$ of group $\bI_j$ using only the class specific feature centers $\bm_c$ in group $\bI_j$ ($c\in\bI_j$). Then, the shared feature center $\bs_{j}$ of the group $\bI_j$ is determined as:
	\begin{equation}
	\bs_j=\frac{\sum_{c\in\bI_j}\bm_c}{C}
	\end{equation}
%	\vspace{-25pt}
	\section{Experiments}
%	\vspace{-5pt}
	Three benchmark {\bf datasets} for fine-grained image classification are used for experiments:
	{\bf CUB-200-2011} \cite{wah2011caltech}: The dataset contains 11788 images of 200 bird species with detailed annotations of parts and bounding boxes. {\bf FGVC-aircraft} \cite{maji2013fine}: The dataset consists of 10000 images across 100 classes denoting a fine-grained set of airplanes with different variants, such as discriminating Boeing 727-300 from Boeing 737-400. 
	\begin{figure}
		\begin{tabular}{ccc}
			\centering
			\bmvaHangBox{{\includegraphics[width=3.95cm]{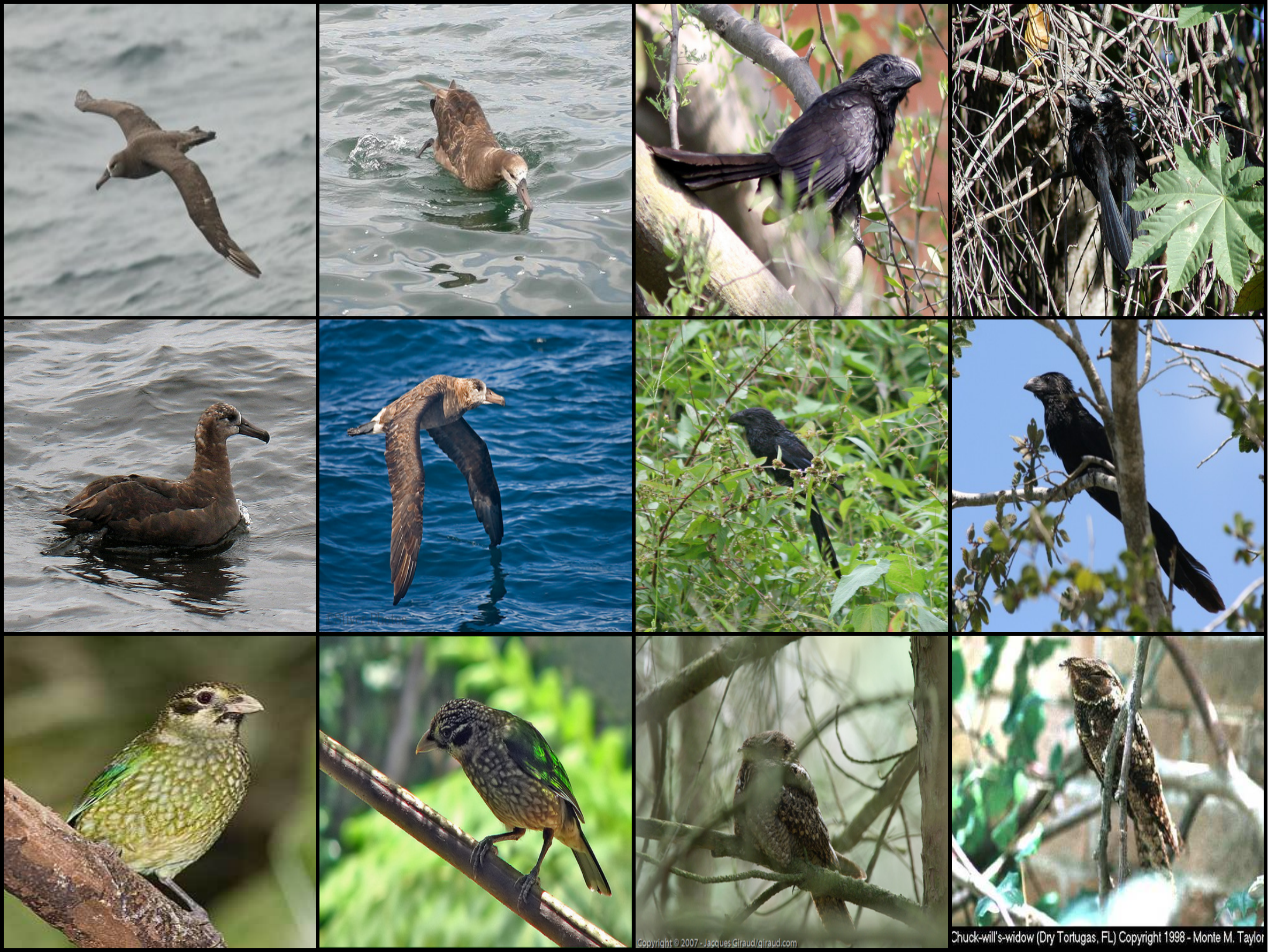}}}&
			\bmvaHangBox{{\includegraphics[width=3.95cm]{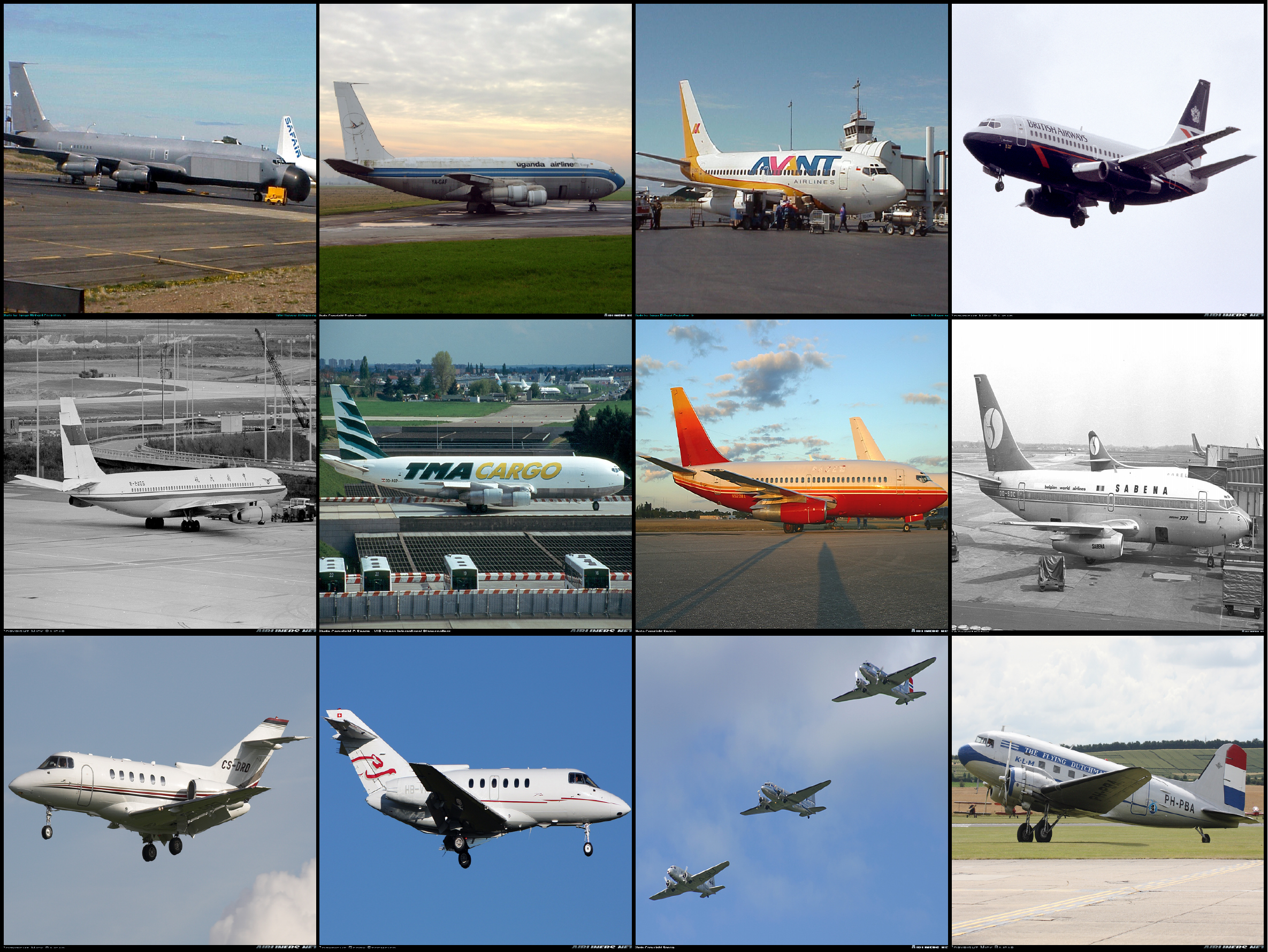}}}&
			\bmvaHangBox{{\includegraphics[width=3.95cm]{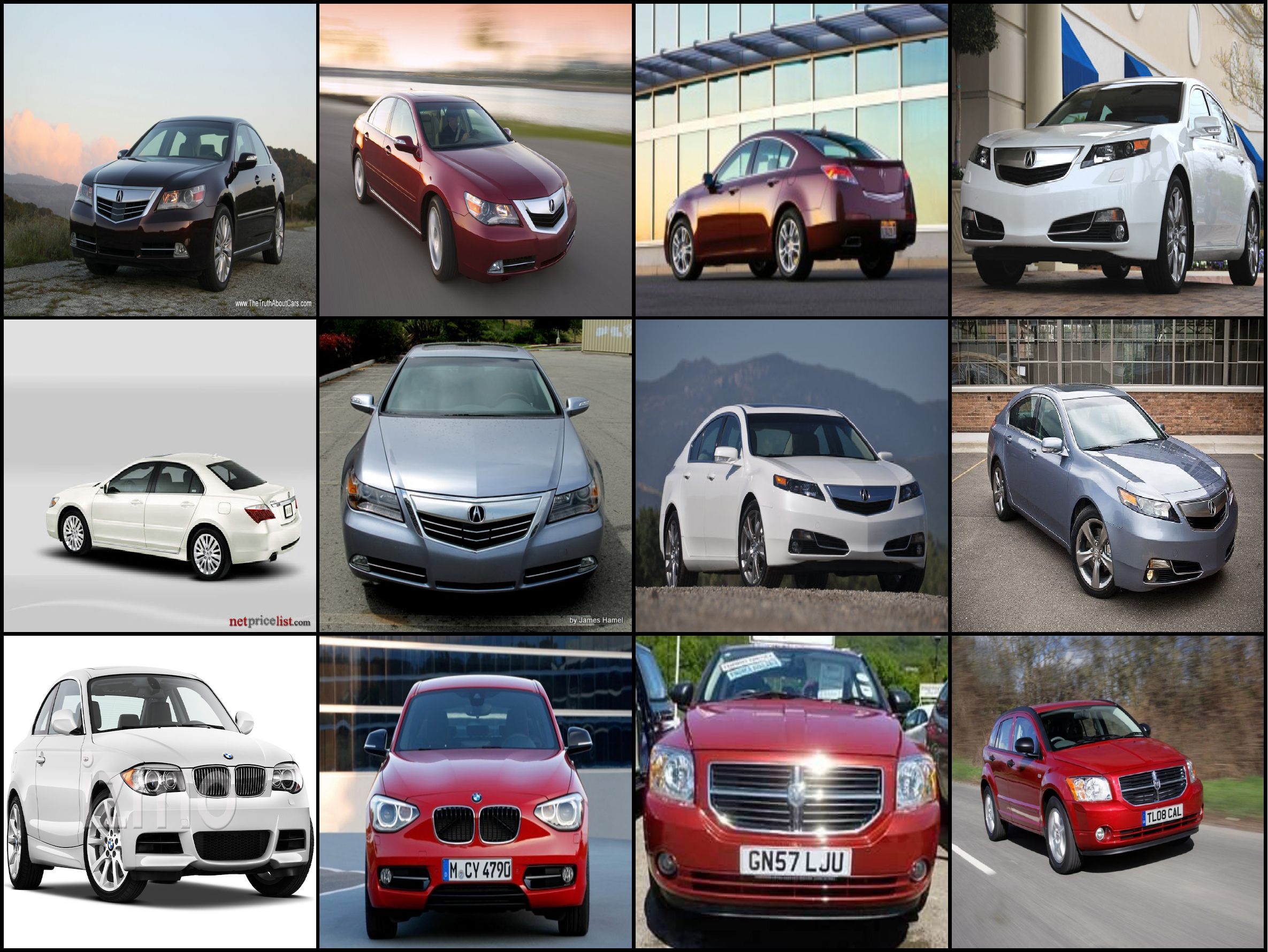}}}\\
			{\footnotesize (a) CUB-200-2011}&{\footnotesize (b) FGVC-aircraft}&{\footnotesize (c) Stanford cars}
		\end{tabular}
		\caption{The image samples from the three datasets: the objects in the images share much similarity but have lots variations in poses and surrounding environments}
		\label{fig:ImageExample}
%		\vspace{-10pt}
	\end{figure}
	{\bf Stanford cars} \cite{krause20133d}: This dataset contains 16185 images of 196 classes. The image samples of these three datasets are shown in Fig. \ref{fig:ImageExample}. For convenience, the three datasets are denoted as "birds", "aircrafts" and "cars" in the following experiments. The training-test split for the three datasets are consistent with what is reported in their work \cite{wah2011caltech,maji2013fine,krause20133d} and used widely in other comparisons. 
	%All the state-of-the art algorithms we are going to compare with have been successfully applied to these datasets under the same train-test setting. Therefore, we reported the corresponding results of each dataset from the original papers directly in the following experiments for the fair comparison.
	
	{\bf Training details:} For the experiments where VGG 16 \cite{simonyan2014very} is directly used for feature extraction, the parameters are set as below: input image size: 224$\times$224, optimizer: Adam, learning rate of 0.0001, which reduces by a factor of 0.6 every 10 epoch after 20 epochs, weight decay rate of 0.0001, and the values of $\alpha_1$, $\alpha_2$, $\alpha_3$, and $\eta$ are set as 0.01 according to the results of cross-validation \cite{monga2017handbook}. The only data augmentation we employ is the horizontal flip. For the experiments where VGG 16 is not directly used for feature extraction (see Section \ref{subsec:improvements}), the experimental setting follows standard practices in fine-grained image classification. Namely, the image input size is 448$\times$448, and it is obtained by resizing the original image so that the shorter side is 448 while keeping its aspect ratio. Then a square image of size 448$\times$448 is cropped from the resized image. For the sake of faster convergence and better performance, we first train the whole network without fine-tuning the weights in the feature extraction part. Then, we use the obtained weights as the initial weights for the classification part and fine tune the whole network. The other parameters are set as below: optimizer: Adam, learning rate: 0.001 for transfer learning (it will times 0.1 if the accuracy does not increase for continuous 6 epochs), and 0.0001 for fine-tuning (reduces by a factor of 0.65 if the accuracy does not increase for continuous 6 epochs), weight decay rate: 0.00001, the values of $\alpha_1$ and $\eta$ are set as 0.01, and $\alpha_2$, $\alpha_3$ are set as 0.1. 
	%For fair comparison, in Section \ref{subsubsec:JDSFLimprovements}, except for above hyper parameters and procedures, we try our best to follow other experimental details mentioned in \cite{gao2016compact} and \cite{cui2017kernel} to ensure the results are valid.
	
	It is noted that in Section \ref{subsec:ablation}, all the experiments are conducted using the pre-trained VGG 16 model as the feature extraction model for comparison convenience.
%	\vspace{-10pt}
	\subsection{Ablation study}\label{subsec:ablation} 
%	\vspace{-3pt}
	\subsubsection{Benefits of Group Based Learning}
	\label{subsubsec:ClassNumber}
%	\vspace{-3pt}
	\begin{figure*}[t]
		\centering
		\includegraphics[width=1\textwidth]{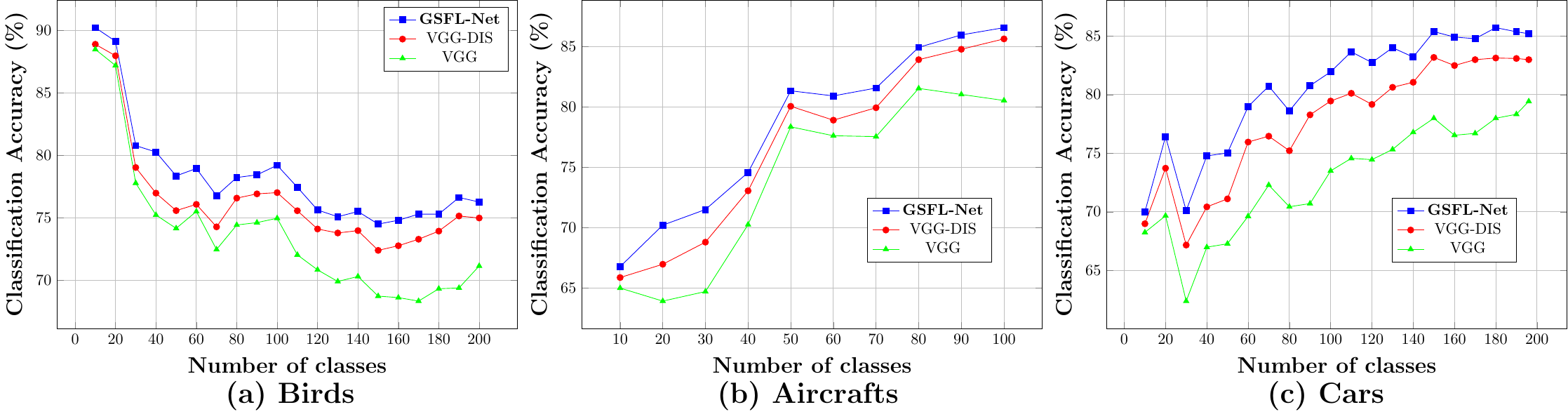}
%		\vspace{-15pt}
		\caption{Classification accuracy w.r.t. different number of classes ($N_g=1$)}\label{fig:CategoryNumber}
%		\vspace{-5pt}
	\end{figure*}
	We investigate the reason of learning multiple shared feature patterns by comparing the classification accuracy change w.r.t. the change of the number of classes using three different methods: fine-tuning VGG 16 network (denoted as VGG), VGG-DIS (structure without Encoder.S and Decoder, and $\alpha_1=\alpha_2=0$) and GSFL-Net by setting $N_g=1$. The number of classes is increased by adding more classes to the base of the previous one. The final results are shown in Fig.\ref{fig:CategoryNumber}. It is found that GSFL-Net always performs best under all the conditions, VGG-DIS ranked second place and VGG performs worst. As the number of classes increases, the advantages of GSFL-Net and VGG-DIS become more and more obvious compared with VGG. However, the advantage of GSFL-Net compared with VGG-DIS becomes relatively smaller as the number of classes becomes larger. This is because as the number of classes goes up, it is harder to find a shared feature center where all the classes have a contribution to it. 
	\begin{table*}[ht]
		\vspace{-16pt} 
		\caption{Epoch numbers required for convergence} % title of Table
		\begin{adjustbox}{width=1\textwidth}
			\centering % used for centering table
			\begin{tabular}{l |c c c c|c c c|c c c c} % centered columns (4 columns)
				\hline\hline %inserts double horizontal lines
				\multirow{3}{*}{Method} & \multicolumn{11}{c}{Convergence epochs}\\ % inserts table
				\cline{2-12}
				& \multicolumn{4}{c|}{Birds} & \multicolumn{3}{c|}{Aircrafts} & \multicolumn{4}{c}{Cars}\\
				%heading
				\hline % inserts single horizontal line
				&  C$10$-$50$ & C$60$-$100$ &  C$110$-$150$ & C$160$-$200$&  C$10$-$30$ & C$40$-$60$ &  C$70$-$100$ &  C$10$-$50$ & C$60$-$100$ &  C$110$-$150$ & C$160$-$200$\\
				%heading
				\hline % inserts single horizontal line
				JDSFL & $<50$ & $50-70$ & $50-70$ & $70-80$& $60-90$ & $60-80$ & $70-90$ & $30-70$ & $70-80$ & $80-100$ & $90-110$\\ % inserting body of the table
				
				VGG-DIS & $<50$ & $50-80$ & $80-90$& $>90$& $60-90$ & $70-90$ & $80-100$& $30-70$ & $70-90$ & $90-110$& $100-110$\\
				
				VGG & $<50$ & $50-80$ & $80-120$& $>120$& $90-100$ & $90-100$ & $90-120$& $30-70$ & $90-100$ & $100-120$& $110-130$\\
				\hline %inserts single line
			\end{tabular}
		\end{adjustbox}
		\label{table:ConvergenceEpoch}
%		\vspace{-30pt}
	\end{table*}
	
	At the same time, we also provide the averaged epoch number required for each method to converge to their best performance as the number of classes increases in Table \ref{table:ConvergenceEpoch}, where we can find that VGG takes more epochs to converge. Meanwhile, GSFL-Net takes fewest epochs to converge compared with the other two methods. This may be due to the reason that during the learning process, by removing the influence of shared components from the features, the rest of the discriminative components are more compactly concentrated around the class-specific feature centers and therefore it is easier for the discriminative components to approximate the class specific feature center. 
	
	Next, we compare the results obtained before and after clustering the classes into different groups for learning multiple shared feature centers for all three datasets. The results of the experiments are shown in Table \ref{table:cluster}. The first two rows show the classification accuracies of VGG-DIS and GSFL-Net when learning only one shared feature center ($N_g=1$). The second two rows show the classification accuracies of GSFL-Net when clustering all the classes into 5 groups ($N_g=5$) either by randomly and uniformly selecting classes for each group or taking the k-means method as mentioned in Section \ref{subsec:updataM} to select classes for each group.
	\begin{table}
		\vspace{5pt}
		\caption{GSFL-Net performance w.r.t. number of groups} % title of Table
		\centering % used for centering table
		\begin{adjustbox}{width=0.78\textwidth}
			\scalebox{1}{
				\begin{tabular}{c l|c| c| c} % centered columns (4 columns)
					\hline\hline %inserts double horizontal lines
					\multicolumn{2}{c}{\multirow{2}{*}{Conditions}}  & \multicolumn{3}{|c}{Classification accuracy($\%$)}\\    % inserts table
					%heading
					\cline{3-5}
					& &Birds (C=200)&Aircrafts (C=100)&Cars (C=196)\\
					\hline % inserts single horizontal line
					
					& VGG-DIS  & $74.99$ &
					$85.63$ & $82.98$\\ % inserting body of the table
					\hline
					$N_g=1$:&GSFL-Net  &$76.27$ & $86.56$ & $85.19$\\
					\hline
					\multirow{2}{*}{$N_g=5$:}& GSFL-Net (Random) & $76.34$& $86.71$ & $85.47$\\
					
					&GSFL-Net ($k$-means)& ${\bf 76.69}$&${\bf 87.07}$ &${\bf 85.83}$\\
					\hline %inserts single line
				\end{tabular}}
			\end{adjustbox}
			\label{table:cluster} % is used to refer this table in the text
%			\vspace{-11pt}
		\end{table}
		The group number $N_g$ is chosen as 5 according to the results of cross-validation \cite{monga2017handbook}.
		We can find that results obtained by using the $k$-means method are the best, and the results obtained by random clustering rank second place. The performance benefits of GSFL-Net (with $N_g = 5$) over VGG-DIS are readily apparent in Table \ref{table:cluster}. This clearly proves that as the number of classes becomes large, learning multiple shared feature centers is more effective than learning only one shared feature center. Furthermore, if the multiple shared feature centers can be learned by classifying classes into groups where classes enjoy lots of similarities, the classification performance can be significantly improved. We use the same settings as for GSFL-Net ($k$-means) in Table \ref{table:cluster} for the following experiments.
%		\vspace{-10pt}
		\subsubsection{Effect of the Feature Expression Loss}
		\label{subsubsec:regularizer}
%		\vspace{-2pt}
		The effects of different configurations of Feature Expression Loss are investigated in this section. Table \ref{table:regularizer} reports the results of our proposed method with different combinations of individual loss terms shown in Eq. \eqref{eq:FE}. From Table \ref{table:regularizer}, it can be found that the classification accuracy of GSFL-Net will drop 1\% to 3\% depending on the what of loss terms utilized.
		\begin{table}[ht]
%			\vspace{-10pt}
			\caption{Different configurations of the loss function and their performances} % title of Table
			\centering
			\begin{adjustbox}{width=1\textwidth}
				\centering % used for centering table
				\begin{tabular}{c|c|c|c|c|c|c|c|c|c|c|c|c} % centered columns (4 columns)
					\hline\hline %inserts double horizontal lines
					\multirow{1}{*}{Loss terms} & \multicolumn{4}{c|}{Birds}&\multicolumn{4}{c|}{Aircrafts}&\multicolumn{4}{c}{Cars}\\    % inserts table
					%heading
					\cline{2-5}
					\hline % inserts single horizontal line
					
					$\alpha_1$\tnote{a}  &   & $\checkmark$ & $\checkmark$& $\checkmark$ &   & $\checkmark$ & $\checkmark$& $\checkmark$  &  & $\checkmark$& $\checkmark$&$\checkmark$\\ % inserting body of the table
					$\alpha_2$& $\checkmark$&  & $\checkmark$&$\checkmark$& $\checkmark$&  & $\checkmark$&$\checkmark$& $\checkmark$&  & $\checkmark$&$\checkmark$\\
					
					$\alpha_3$ & $\checkmark$&$\checkmark$& &$\checkmark$& $\checkmark$&$\checkmark$& &$\checkmark$& $\checkmark$&$\checkmark$& &$\checkmark$\\
					\hline %inserts single line
					$Acc.(\%)$ & $75.88$&$72.51$ &$75.89$&${\bf 76.69}$& $85.59$&$ 84.18$ &$85.68$&${\bf 87.07}$& $84.77$&$ 82.98$ &$84.37$&${\bf 85.83}$\\
					\hline %inserts single line
				\end{tabular}
			\end{adjustbox}
			\begin{tablenotes}
				\item[a]We use the hyperparameters $\alpha_1$, $\alpha_2$ and $\alpha_3$ to represent their corresponding loss terms.
			\end{tablenotes}
			\label{table:regularizer} % is used to refer this table in the text
			\vspace{-15pt}
		\end{table}
%		\vspace{-1pt}
		\subsubsection{Improvements of GSFL-Net in Localization Ability}\label{subsubsec:LocalizationAbility}
%		\vspace{-3pt}
		\begin{figure*}[t]
			\centering
			\includegraphics[width=1\textwidth]{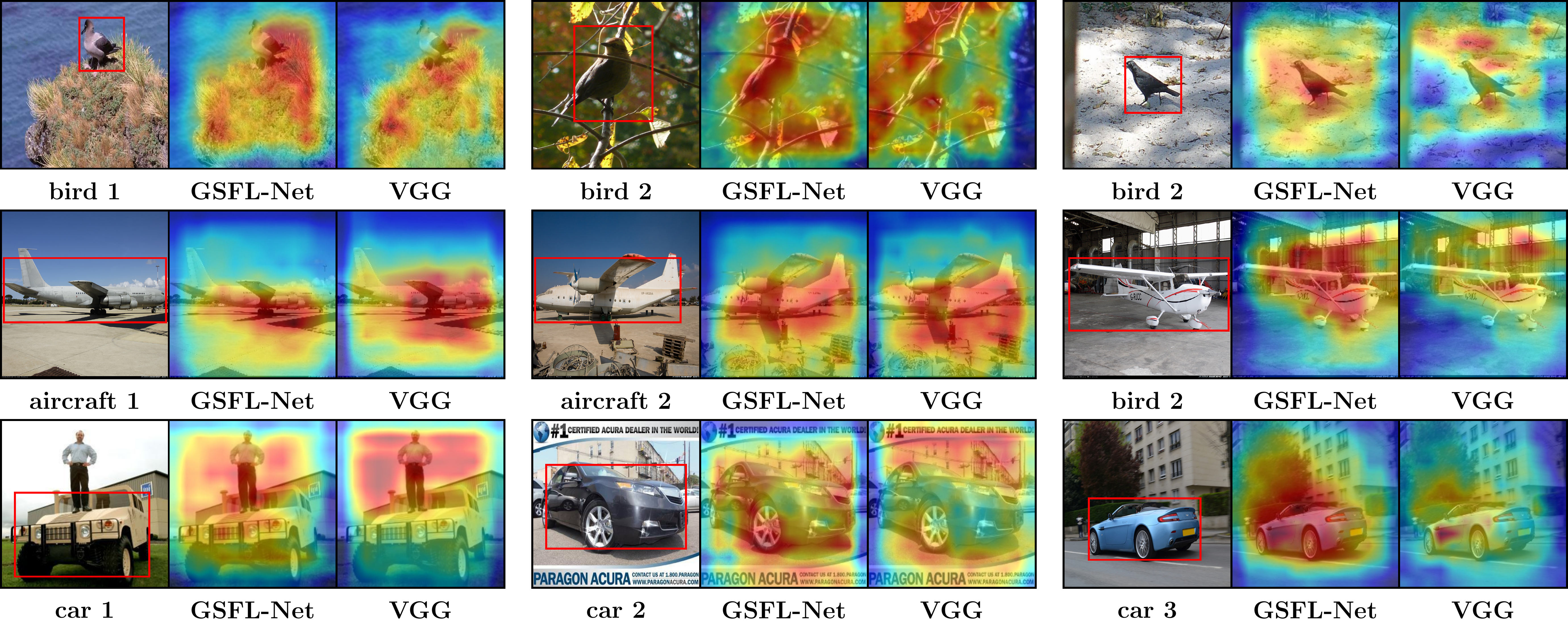}
			\caption{Localization ability improvements brought by GSFL-Net}\label{fig:CradCam}
%			\vspace{-4pt}
		\end{figure*}
		In fine-grained image classification, finding the locations of objects in the images accurately is the key to facilitate the classification work. Hence, we utilize the Gradient-Weighted Class Activation Mapping (Grad-CAM) method \cite{selvaraju2017grad} to investigate whether GSFL-Net improves the localization ability of CNNs in the feature extraction model. Grad-CAM is able to provide a heat map of visual saliency as produced by the CNNs of the network. We can tell from the heat map where the CNNs are focusing on. Fig.\ref{fig:CradCam} shows image samples from the three datasets and their corresponding heat maps produced by using GSFL-Net with VGG 16 as the feature extraction model and fine-tuning the VGG 16 network. The locations of the objects in the images are marked using red boxes. It is easy to find from Fig.\ref{fig:CradCam} that GSFL-Net can better localize the objects in the images since in the heat maps it produces, higher pixel values are seen in areas where target objects are located. %For example, in the heat maps for the bird 2, VGG has missed the most part of the bird body, but GSFL-Net is able to localize the whole body of the bird. In addition, for the aircraft 2, VGG takes the objects below the aircraft as a part of the classification target, on contrary, GSFL-Net does not have the problem. Furthermore, for the car 1, VGG mistakenly regarded the person as the classification target which is needed to be localized. Although GSFL-Net is also distracted by the location of the person in the image, it successfully outlined the shape of the car at the same time. From the results, we can conclude that GSFL-Net does improve the localization ability of the CNNs.
%		\vspace{-15pt}
		\subsection{Comparisons against and improvements to state of the art}\label{subsec:improvements}
		\begin{table}[ht!]
%			\vspace{-18pt}
			\centering
			\caption{Classification accuracy comparison with state-of-the-art methods}
			\begin{adjustbox}{width=0.95\textwidth}
				\centering % used for centering table
				\begin{tabular}{l |c c| c c| c c| c} % centered columns (4 columns)
					\hline\hline %inserts double horizontal lines
					\multirow{2}{*}{Method} &\multicolumn{2}{c|}{Birds($\%$)}&\multicolumn{2}{c|}{Aircrafts($\%$)}&\multicolumn{2}{c|}{Cars($\%$)}&\\
					\cline{2-8}
					& Accuracy & Improvement & Accuracy & Improvement& Accuracy & Improvement&Base Model\\ [0.5ex] 
					%heading
					\hline % inserts single horizontal line
					Lin, T., et al.\cite{lin2018bilinear} & $84.10$ & - & $86.90$ & - & $91.30$ & - & VGG\\ % inserting body of the table
					Shu, K., et al.\cite{kong2017low} & $84.21$ & - & $87.31$ & - & $90.92$ & - & VGG\\ % inserting body of the table
					Wei, X., et al. \cite{wei2018grassmann} & $85.40$ & - & $88.10$ & - & $91.70$ & - & VGG\\
					Zheng, H., et al.\cite{zheng2017learning} & $85.40$ & -& $88.40$ & - & $92.80$ & -  & VGG\\
					Du, A., et al.\cite{dubey2018pairwise} &  $86.87$ & - & $89.24$ & - & $92.86$ & - & DenseNet\\
					\hline
					VGG & $70.40$ &\multirow{2}{*}{($6.29$)}& $80.53$&\multirow{2}{*}{($6.54$)}& $79.42$&\multirow{2}{*}{($6.41$)}&\multirow{2}{*}{VGG}\\
					GSFL-Net(VGG) & $76.69$ & & $87.07$ & & $85.83$ & &\\
					\hline
					Compact Bilinear CNN\cite{lin2018bilinear} & $84.00$ &\multirow{2}{*}{($1.81$)} & $87.10$ &\multirow{2}{*}{($2.16$)} & $90.19$ &\multirow{2}{*}{($2.00$)}&\multirow{2}{*}{VGG}\\
					GSFL-Net(Compact bilinear feature model) & 85.81 && ${\bf 89.26}$ && $92.19$ & &\\
					\hline %inserts single line
					Kernel Bilinear CNN\cite{cui2017kernel} & $86.20$ &\multirow{2}{*}{(1.40)}& $86.90$ &\multirow{2}{*}{($1.33$)}& $92.90$ &\multirow{2}{*}{($1.02$)}&\multirow{2}{*}{VGG}\\
					GSFL-Net(Kernel bilinear feature model) & ${\bf 87.60}$ & &  $88.23$ & & ${\bf 93.92}$ & &\\
					\hline %inserts single line
				\end{tabular}
			\end{adjustbox}
			% title of Table
			\label{table:improvements} % is used to refer this table in the text
%			\vspace{-13pt}
		\end{table}
		Fine-grained image datasets are tough datasets which cannot be easily classified by only using pre-trained models such as VGG 16, Google Net \cite{szegedy2015going}, or ResNet \cite{he2016deep} directly without enough object annotations. Although DenseNet may be effective as shown in \cite{dubey2018pairwise}, without enough GPUs, it is hard to make modifications on DenseNet due to its large number of layers and high-dimensional extracted features. Fortunately, as described in Section \ref{sec:intro}, our GSFL-Net can be combined with any feature extraction model and trained from end to end to improve their performance. In this Section, we select several representative feature extraction models which have been successfully applied to all three fine-grained datasets investigated in this paper. 
		
		The feature extraction models we employ are: 1.) the baseline VGG 16 model, 2.) the compact bilinear model \cite{gao2016compact}, which can learn features that have the same representative power compared with full bilinear representations but with much fewer dimensions, and 3.) the kernel pooling method in \cite{cui2017kernel}, which has a feature extraction model that is able to capture higher order interactions of features in the form of kernels. The results shown in Table \ref{table:improvements} verify that by combining our GSFL-Net with known feature extractors can improve their performance by $1-6\%$. Furthermore, it can also be found that the simpler (and weaker) the original feature extraction model is, the larger the improvement will be. Indeed, the GSFL-Net variants in Table \ref{table:improvements} help achieve a new state of the art performance. It is worth noting that  GSFL-Net can also be combined with their feature extraction models proposed in \cite{lin2018bilinear,wei2018grassmann,yu2018hierarchical,kong2017low,zheng2017learning}.
%		\vspace{-20pt}
		\section{Conclusion}
		\label{sec:conclusion}
%		\vspace{-10pt}
		We present a new deep network, GSFL-Net, that explicitly models shared features and removes their effect to enhance fine-grained image classification accuracy. Our modeling of shared features is based on a new group based learning wherein existing classes are divided into groups and multiple shared feature patterns are discovered. In this work, we demonstrated that: a.) multi-group based shared feature learning can outperform using a single group to model all shared features, and 2.) GSFL-Net, when combined with state of the art feature extractors, can improve their performance and further extend achievable state of the art performance on challenging benchmark datasets. 

\bibliography{egbib}
\end{document}